# Optimal Monte Carlo Estimation of Belief Network Inference

**Malcolm Pradhan**
<pradhan@camis.stanford.edu>

**Paul Dagum**
<dagum@camis.stanford.edu>

Section on Medical Informatics
MSOB, X-215
Stanford University, CA 94305

## Abstract

We present two Monte Carlo sampling algorithms for probabilistic inference that guarantee polynomial-time convergence for a larger class of network than current sampling algorithms provide. These new methods are variants of the known likelihood weighting algorithm. We use of recent advances in the theory of optimal stopping rules for Monte Carlo simulation to obtain an inference approximation with relative error $\varepsilon$ and a small failure probability $\delta$. We present an empirical evaluation of the algorithms which demonstrates their improved performance.

## 1 INTRODUCTION

Exact probabilistic inference in belief networks is known to be NP-hard in the worst case [Cooper, 1990], but even obtaining an exact solution in real-world networks may also be impractical [Pradhan et al., 1994; Shwe et al., 1991]. This complexity result encouraged researchers to explore approximate inference, in particular Monte Carlo simulation and search techniques. The simulation methods include Gibbs sampling (straight sampling) [Pearl, 1987], likelihood weighting [Fung & Chang, 1990; Shachter & Peot, 1990], logic sampling [Henrion, 1988], and randomized approximation schemes [Chavez & Cooper, 1990]. Many variations of these algorithms have been reported that improve on the run times [Fung & Chang, 1990; Fung & Del Favero, 1994; Hulme, 1995; Shachter & Peot, 1990; Shwe & Cooper, 1991].

Dagum and Luby [Dagum & Luby, 1993] showed that the general problem of approximate inference in belief networks with evidence is also NP-hard. There are, however, restricted classes of networks in which approximate inference is provably amenable to a polynomial time solution. [Dagum & Chavez, 1993].

In this paper we present two randomized approximation algorithms, the *bounded variance* algorithm and the $\mathcal{AA}$ algorithm, that make use of recent advances in stopping rules for Monte Carlo sampling [Dagum et al., 1995]. If the belief network does not contain *extreme conditional probabilities* (defined in Section 2.2) then these algorithms can approximate an inference in worst-case polynomial time, but with a small probability the algorithm may fail to output an approximation within specified limits. The class of belief networks that does not contain extreme conditional probabilities is a much larger class than the class studied by Dagum and Chavez [Dagum & Chavez, 1993].

Previous simulation algorithms for probabilistic inference, such as likelihood weighting and logic sampling, are known to require exponential running time to converge to small inference probabilities. In contrast, bounded variance and $\mathcal{AA}$ often approximate these inferences in polynomial time, requiring exponential time only if there are extreme probabilities in the evidence nodes, $E$, or hypothesis nodes, $H$. These new algorithms also have the attractive property of allowing the user to know when approximating an inference requires exponential time computation, allowing a meta-reasoner to trade-off running time with approximation accuracy.

In the next section we review sampling algorithms and highlight common practical problems associated with their use. The bounded variance and $\mathcal{AA}$ algorithms are modifications of the likelihood weighting algorithm, which is reviewed in Section 2.3. In Section 3 we present recent work on stopping rules for Monte Carlo simulation. Section 4 is a description of the bounded variance and $\mathcal{AA}$ algorithms; an empirical evaluation of the new algorithms is presented in Section 5.

## 2 A REVIEW OF SAMPLING ALGORITHMS

The algorithms we present in Section 4 yield *relative approximations* of inference. Before describing the likelihood weighting algorithm we will review different types of approximation algorithms, in particular we make clear the distinction between relative and absolute error bounds.

### 2.1 A CATEGORIZATION OF APPROXIMATION ALGORITHMS

The following discussion is modeled after [Dagum & Luby, 1993].

INSTANCE: A real value $\varepsilon$ between 0 and 1, a belief network with binary valued nodes, $V$, arcs $A$, conditional probabilities Pr, two nodes $X$ and $E$ in $V$ instantiated to $x$ and $e$, respectively.

ABSOLUTE APPROXIMATION: An estimate $0 \leq \phi \leq 1$ such that

$$\Pr[X = x | E = e] - \varepsilon \leq \phi \leq \Pr[X = x | E = e] + \varepsilon.$$

RELATIVE APPROXIMATION: An estimate $0 \leq \phi \leq 1$ such that

$$\Pr[X = x | E = e](1 - \varepsilon) \leq \phi \leq \Pr[X = x | E = e](1 + \varepsilon).$$

In addition, an algorithm is *deterministic* if it guarantees to produce an approximation $\phi$ within the specified bounds. Search based algorithms [Cooper, 1984; Henrion, 1991] are usually of this variety. A *randomized* approximation algorithm produces an approximation $\phi$ within the specified bounds with a small failure probability $\delta > 0$. Sampling algorithms (using random bits) fall into this second category.

Note that Chebychev's inequality proves we can approximate $\Pr[X = x, E = e]$ and $\Pr[E = e]$ with *absolute* errors in polynomial time. In general, however, we cannot use these approximations to estimate $\Pr[X = x | E = e]$ with any type of error. In contrast, relative approximations of $\Pr[X = x, E = e]$ and $\Pr[E = e]$ will yield a relative approximation of $\Pr[X = x | E = e]$.

Monte Carlo sampling methods can be classified as short run or long run algorithms. Short run algorithms, such as logic sampling and likelihood weighting, produce an estimate of an inference probability by randomly generating independent instances and taking the expected value. Long run algorithms, in particular Gibb's sampling, also known as Pearl's straight simulation [Hrycej, 1990; Pearl, 1987], are forms of Markov chain sampling, and therefore will converge to the expected value in the limit if certain properties hold. Error estimation is difficult in Markov chain algorithms because instances are not independent. Since our focus in this paper is the reduction and measurement of error we will concentrate on the short run algorithms.

## 2.2 CHARACTERIZING APPROXIMATION COMPLEXITY

For a given $(\varepsilon, \delta)$ a randomized approximation algorithm has a polynomial running time if it outputs a relative approximation in the size of the network $n$, $\varepsilon^{-1}$, and $\ln \delta^{-1}$.

Dagum and Luby [Dagum & Luby, 1994] defined a *local variance bound* (LVB) of a belief network to describe the range of conditional probabilities contained in a belief network, and complexity of inference. Let $\Gamma$ be the LVB of a binary valued network,

$$\Gamma = \max\left[\frac{u}{l}, \frac{1 - l}{1 - u}\right]$$

where l and u are real numbers in [0,1], such that $l \leq u$, and all conditional probabilities $\Pr[X = 0 | \pi(X)]$ for each node $X$ in a belief network are contained in either [l,u] or [1-u,1-l]. The class of networks of size $n$ is said to contain *extreme conditional probabilities* if the LVB is not bounded by $n^c$ for some integer $c > 0$.

As described the LVB is useful for characterizing a network, but for Monte Carlo simulation of inference we are only concerned with the LVB of a subset, $m$, of the network consisting of the evidence nodes, and the nodes we will query (the hypothesis nodes). If the LVB of this subset is bounded by $n^c$ then the approximation will be in polynomial time, otherwise it is NP-hard [Dagum & Luby, 1993]. The practical implication of this fact is that if our network contains extreme conditional probabilities we cannot say a priori how many iterations to run our sampling algorithm because it will depend on the LVB of evidence set, and the nodes to be queried. In Section 3 we show how statistical stopping rules help us overcome this problem.

## 2.3 LIKELIHOOD WEIGHTING

The likelihood weighting algorithm [Fung & Chang, 1990; Shachter & Peot, 1990] has been the most implemented Monte Carlo simulation methods used for belief network inference, in part because of it's ease of implementation and faster convergence times compared to logic sampling [Cousins et al., 1993; Shachter & Peot, 1990]. In this section we review the likelihood weighting algorithm since it forms the basis for the algorithms presented in this paper.

In the following discussion, let $E$ denote the set of observed nodes of a belief network, and $Z$ the nodes not contained in $E$. The set of parents of a node $X_i$ is represented by $\pi(X_i)$. Lowercase letters denote a particular instantiation of the variables. Expressions are conditioned, for example $\prod_{Z_i \in Z} f(Z_i)|_{Z = z, E = e}$, to denote an instantiation of their arguments, in this case $Z$ to $z$, and $E$ to $e$.

The basic likelihood weighting algorithm orders the nodes in the belief network in parent ordering and assigns each node in $Z$ to a state value. Since evidence nodes $E$ are already set, this process results in an network *instance*, $(z, e)$ (Figure 1). A *path probability* is scored:

$$\rho(z, e) = \prod_{Z_i \in Z} \Pr[Z_i | \pi(Z_i)]\Big|_{Z = z, E = e}.$$

Only the unobserved nodes, $Z$, are sampled hence the path probability differs from the full joint probability of the belief network. In the literature the path probability has been termed the *probability of selecting* the instance.

A weighting distribution $\omega(z, e)$ is used to obtain an unbiased score:

$$\omega(z, e) = \prod_{E_i \in E} \Pr[E_i | \pi(E_i)]\Big|_{Z = z, E = e}.$$

An indicator, $\chi(z, e)$, is 1 if $Z = z$ instantiates the node $X$ to $x$, or 0 otherwise. Likelihood weighting estimates $\Pr[E = e]$ with $\omega(z, e)$, and $\Pr[X = x, E = e]$ with $\chi(z, e) \cdot \rho(z, e) \cdot \omega(z, e)$.

A common implementation of the standard likelihood weighting algorithm involves "binning" to score the network in the $\chi(z, e) \cdot \omega(z, e)$ step, followed by renormalization to obtain probability estimates [Shachter & Peot, 1990]. For example, consider a binary query node $X_i$ for which we are interested in an estimate of $\Pr[X_i = 1 | E = e]$. If the prior probability, $\Pr[X_i = 1]$, is small then almost all of our samples will be generated with $X_i = 0$. Very few samples will score $X_i = 1$, however, when renormalizing after scoring we obtain an estimate on $\Pr[X_i = 1 | E = e]$ for which we cannot estimate a relative error since we have used information from estimation of the complement state.

Likelihood weighting and logic sampling [Henrion, 1988] are example of simulation algorithm that estimate $\Pr[X = x | E = e]$ from the ratios of $\Pr[X = x, E = e]$ and $\Pr[E = e]$. These algorithms require exponential time for rare hypothesis ($\Pr[X = x]$), rare evidence ($\Pr[E = e]$), or rare inference ($\Pr[X = x | E = e]$) configurations. Consider a belief network that contains an exponentially small probability in the conditional probability $\Pr[X = x | \pi(X)]$, the likelihood weighting algorithm must generate an exponential number of samples before it samples an instance consistent with the query state, $X = x$.

Even in situations where a network does not contain any extreme probabilities, previous approximate inference algorithms may not reliably produce estimates in polynomial time. This problem is further aggravated by the lack of an error guarantee $\varepsilon$ and a failure guarantee $\delta$ by these algorithms.

## 3    STOPPING RULES

We next discuss stopping criteria, that is, how many samples are required before the algorithm achieves the specified $(\varepsilon, \delta)$-estimation. Powerful stopping criteria allow us to develop powerful approximation algorithms.

Chebychev's inequality lets us define a distribution-independent upper bound on $N$, the number of trials to run the simulation. Let $Y$ be a random variable in [0,1] with mean $\phi$ and variance $\sigma^2$, the true value is $\mu$:

$$\Pr\left[ \left| \frac{\phi - \mu}{\mu} \right| > \varepsilon \right] < \frac{c\sigma^2}{N\varepsilon^2\mu^2}$$

for some small constant, $c$. The probability that the relative error exceeds $\varepsilon$ is the failure probability $\delta$. We can rearrange the inequality to estimate the number of samples $N$:

$$N \geq c \cdot \frac{\sigma^2}{\varepsilon^2\mu^2} \cdot \frac{1}{\delta}. \tag{1}$$

Note that since $\sigma^2 = \mu(1 - \mu)$ for a Bernoulli random variable, we get that

$$N \geq \frac{c}{\mu\varepsilon^2} \cdot \frac{1}{\delta}.$$

Tighter bounds can be estimated using Zero-One Estimation Theory [Karp et al., 1989] which produces a lower bound on the number of Bernoulli trials required to achieve a specified level of accuracy. Monte Carlo simulation for inference can be framed as a series of Bernoulli trials where success is defined as simulating an instance which contains a variable of interest $X_1 = x$ [Dagum &

Horvitz, 1993]. Zero-One Estimation Theory gives the following result for the upper bound on $N$:

$$N = \frac{4}{\mu\varepsilon^2} \ln\frac{2}{\delta}. \tag{2}$$

The problem with the estimates given in Eq. 1 and Eq. 2 is that the quantity we are estimating, $\mu$, is required in the calculation for $N$. Dagum and Horvitz [Dagum & Horvitz, 1993] use a Bayesian approach to define a conjugate distribution over the parameter of estimation to circumvent this limitation.

A non-Bayesian method used to remove the dependence of the number of trials $N$ to the parameter of estimation is to use a bound on the parameter. The LVB can be used to bound $\mu$. Let $W$ be the set of nodes in the network that includes the observed variables $E$ and nodes $X$ to be queried, and $k = |W|$. Using the LVB the Zero-One Estimator Theorem (Eq. 2) can be rewritten

$$N \leq \frac{4\Gamma^k}{\varepsilon^2} \ln\frac{2}{\delta} \tag{3}$$

where $N$ is the number of samples to approximate $\mu$ with relative error $\varepsilon$. Unfortunately, the estimate on the number of samples using Eq. 3 is a conservative worst case estimation, and convergence may occur in a much smaller number of trials, as Dagum and Luby [Dagum & Luby, 1994] prove in their paper, and we discuss in Section 4.1.

Both the Bayesian and LVB methods for calculating the number of samples required for a $(\varepsilon, \delta)$ estimation are suboptimal because the Zero-One Estimator Theorem assumes trials of zero or one, which results in a larger variance estimation than actually occurs since our random variables lie in the interval of [0,1]. As shown by Eq. 1, variance reduction results in faster convergence.

In the next section we describe the new algorithms and the techniques used to avoid some of the limitations of stopping rules based on the Zero-One Estimator Theorem.

## 4    IMPROVED ALGORITHMS

Both bounded variance and the $\mathcal{A}\mathcal{A}$ algorithm are modifications of the likelihood weighting algorithm. Faster convergence is achieved by reducing the variance of sample estimates by avoiding the use of an indicator variable, $\chi(z, e)$, to score $\Pr[X = x, E = e]$.

In addition to the improved convergence, the algorithms use several advances in stopping rules for Monte Carlo simulation [Dagum et al., 1995]:

1.  A generalized form of the Zero-One Estimator Theorem for random variables in the interval [0,1].

2.  A stopping rule theorem is described that represents the number of samples required as a random variable.

3.  Sequential analysis methods are used in the $\mathcal{A}\mathcal{A}$ algorithm to better estimate the parameter of interest rather than using uninformative bounds.

## 4.1  THE BOUNDED VARIANCE ALGORITHM

To avoid the use of an indicator variable for scoring $\Pr[X = x, E = e]$, each query node $X_i$ is set to its query state and $\Pr[X = x, E = e]$ is estimated directly. This means that the new algorithms are query specific—they specifically estimate the posterior probability of a set of query nodes $X$. Let the set $W = E \cup \{X_i = x_i\}$, where $X_i$ is a query node set to its query state $x_i$. The likelihood weight score for the network is

$$\omega(z, w) = \prod \Pr[W_i = w_i | \pi(W_i)] \big|_{W = w, Z = z}. \quad (4)$$

For belief networks with extreme conditional probabilities the bounded variance algorithm can renormalize the outcomes of the random variables used to estimate $\Pr[E = e]$ and $\Pr[X = x, E = e]$ such that it converges in polynomial time even if $\Pr[X = x | E = e]$ is exponentially small.

Unlike the stopping rules presented in Section 3, the bounded variance algorithm uses a generalized form of the

---

**Inputs:** $0 < \varepsilon \le 2$, $\delta > 0$, query nodes $X_i$, $i = 1, \dots, n$, evidence nodes $E_j$, $j = 1, \dots, k$

**Bounded Variance:**
$\lambda = (e - 2) \approx 0.72$, $S^* \leftarrow 4\lambda \ln(2/\delta)(1 + \varepsilon)/\varepsilon^2$,
$T_E \leftarrow 0$, $\Pi_E \leftarrow \prod_{j=1}^{k} u_{E_j}$
**for** $i \leftarrow 1$ **to** $n$
$\quad T_i \leftarrow 0$, $\mathcal{W}_i \leftarrow E \cup \{X_i = x_i\}$, $S_i \leftarrow 0$
$\quad \Pi_i \leftarrow u_{X_i} \cdot \Pi_E$

**while** $\exists S_i < S^*$, $i = 1, \dots, n$
$\quad$ **if** $S_E < S^*$ **then**
$\quad\quad z_E \leftarrow$ Generate Instance$(Z, E)$
$\quad\quad \omega_E \leftarrow \prod_{j=1}^{k} \Pr[E_j = e_j | \pi(E_i)] \big|_{E = e, Z = z_E}$
$\quad\quad S_E \leftarrow S_E + \omega_E / \Pi_E$
$\quad\quad T_E \leftarrow T_E + 1$

$\quad$ **for** $i \leftarrow 1$ **to** $n$
$\quad\quad$ **if** $S_i < S^*$ **then**
$\quad\quad\quad z_{\mathcal{W}_i} \leftarrow$ Generate Instance $(Z \backslash X_i, \mathcal{W}_i)$
$\quad\quad\quad \omega_i \leftarrow \prod_{j=1}^{k} \Pr[W_j = w_j | \pi(W_j)] \big|_{\mathcal{W}_i = w, Z = z_{\mathcal{W}_i}}$
$\quad\quad\quad S_i \leftarrow S_i + \omega_i / \Pi_i$
$\quad\quad\quad T_i \leftarrow T_i + 1$

$\phi_E \leftarrow \Pi_E S_E / T_E$
**for** $i \leftarrow 1$ **to** $n$
$\quad \phi_i = \Pi_i S_i / T_i$
$\quad \hat{\mu}_i \leftarrow \phi_i / \phi_E$

**Output:** $\hat{\mu}_i$, $\phi_i$, $i = 1, \dots, n$

**Figure 1.** The bounded variance algorithm

---

Zero-One Estimator Theorem for random variables in the interval $[0,1]$. Instead of defining the number of *successes* for a zero-one random variable, let $S_t = \zeta_1 + \dots + \zeta_t$ be the sum of independent identically distributed random variables, then define $S_t \ge k$ and $S_t < k + 1$ to contain $k$ successes.

The Zero-One Estimator Theorem (Eq. 3) is commonly used to estimate the number of instances required to achieve a $(\varepsilon, \delta)$ inference estimate. The theorem as stated uses the LVB to determine the worst-case variance in the network. Rather than estimating the target number of samples $N$ a priori, bounded variance uses a stopping rule to avoid using the LVB.

Scores are weighted by the upper bound on the variance of the variables being scored, $W$. We define the interval $[l_i, u_i]$ to contain each conditional probability $\Pr[W_i = w_i | \pi(W_i)] \big|_{W = w, Z = z}$, and we form a new random variable:

$$\zeta(z, w) = \frac{\omega(z, w)}{\prod_{i=1}^{k} u_i}. \quad (5)$$

Each sample iteration generates an instance $z_1, z_2, \dots$ of the unobserved nodes $Z$, we denote the sum of the first $t$ samples

$$S_t = \zeta(z_1, w) + \dots + \zeta(z_t, w).$$

We define $\lambda = (e - 2) \approx 0.72$. The algorithm is run until

$$S_T \ge \frac{4\lambda(1 + \varepsilon)}{\varepsilon^2} \ln \frac{2}{\delta} \quad (6)$$

where $T$ is the number of samples. The estimate

$$\phi = \frac{S_T}{T} \prod_{i=1}^{k} u_i$$

is guaranteed to be within a relative error $\varepsilon$ with probability at least $(1 - \delta)$ [Dagum & Luby, 1994]. Eq. 6 represents the *Stopping Rule Theorem*, adapted from [Dagum et al., 1995]. Thus the bounded variance algorithm moves towards the stopping criteria (Eq. 6) for each query node.

Bounded variance will usually stop after far fewer iterations than the worst-case estimate given by the Zero-One Estimator Theorem (Eq. 3) since the actual variance in the set $W$ will usually be less than an estimate based on the LVB.

In the pseudo-code shown in Figure 1 it can be seen that for each query node, $X_i$, a new instance with $X_i = x_i$ must be generated and scored. This extra cost associated with bounded variance can be expensive if $X$ is large. A technique for reducing the cost of regenerating instances is discussed in Section 4.4.

## 4.2  THE $\mathcal{AA}$ ALGORITHM

The $\mathcal{AA}$ algorithm is an implementation of the optimal approximation technique described in [Dagum et al., 1995], and is presented here without proof. Pseudo-code

**Inputs:** $0 < \varepsilon \le 2$, $\delta > 0$, query nodes $X_i$, $i = 1, ..., n$, evidence nodes $E_i$, $i = 1, ..., k$.

$\mathcal{AA}$ **Step 1:**
$\varepsilon' \leftarrow 1/2$, $\delta' \leftarrow \delta/3$, $\lambda = (e-2) \approx 0.72$, $c \leftarrow 2$
$\phi_i$, $i = 1, ..., n \leftarrow$ Bounded variance$(\varepsilon', \delta', X, E)$

$\mathcal{AA}$ **Step 2:**
$\Upsilon = c4\lambda \ln(2/\delta)/\varepsilon^2$
$N_E \leftarrow \Upsilon \cdot \varepsilon/\phi_E$
for $i \leftarrow 1$ to $n$
$\quad N_i \leftarrow \Upsilon \cdot \varepsilon/\phi_i$, $a_i \leftarrow 0$, $\mathcal{W}_i \leftarrow E \cup \{X_i = x_i\}$

for $j \leftarrow 1$ to max$[N]$
$\quad$ **if** $j < N_E$ **then**
$\qquad z_E \leftarrow$ GenerateInstance$(Z, E)$
$\qquad \omega_E \leftarrow \prod_{j=1}^k \Pr[E_j = e_i | \pi(E_i)]|_{E = e, Z = z_E}$
$\qquad$ **if** odd$[j]$ **then** $a_E' \leftarrow \omega_E$
$\qquad\qquad$ **else** $a_E \leftarrow (a_E' + \omega_E)^2/2$

$\quad$ for $i \leftarrow 1$ to $n$
$\qquad$ **if** $j < N_i$ **then**
$\qquad\qquad z_{\mathcal{W}_i} \leftarrow$ Generate Instance $(Z \backslash X_i, \mathcal{W}_i)$
$\qquad\qquad \omega_i \leftarrow \prod_{j=1}^k \Pr[W_j = w_j | \pi(W_j)]|_{\mathcal{W}_i = w, Z = z_{\mathcal{W}_i}}$
$\qquad\qquad$ **if** odd$[j]$ **then** $a_i' \leftarrow \omega_i$
$\qquad\qquad\qquad$ **else** $a_i \leftarrow (a_i' + \omega_i)^2/2$

$\hat{\rho}_E \leftarrow \max[a_E/N_E, \varepsilon\phi_E]$
for $i \leftarrow 1$ to $n$
$\quad \hat{\rho}_i \leftarrow \max[a_i/N_i, \varepsilon\phi_i]$
*continued in step 3...*

**Figure 2.** $\mathcal{AA}$ algorithm steps 1 and 2

for the algorithm is given in Figure 2 and Figure 3. The $\mathcal{AA}$ algorithm attempts to find a smaller bound for the estimate of $\mu$ by using a three step approach, and thereby derive a tighter stopping rule. Let $Y$ be a random variable in [0,1] with mean $\mu$ and variance $\sigma^2$. Let $Y_1, Y_2, ...$ be independently and identically distributed according to $Y$. To achieve a relative error $\varepsilon$ with probability $(1 - \delta)$

1. The bounded variance algorithm (Stopping Rule Algorithm) is run with inputs $\varepsilon' = 1/2$, $\delta' = \delta/3$. This produces quick estimate $\hat{\mu}$ of $\mu$. Let

$$\Upsilon = c4\lambda \ln(2/\delta)/\varepsilon^2$$

for some constant $c > 1$.

2. To estimate the variance of $\hat{\mu}$, $\hat{\rho}$, set $N = \Upsilon \cdot \varepsilon/\hat{\mu}$ and initialize $a \leftarrow 0$. For $i = 1, ..., N$ incrementally calculate the variance:

$$a \leftarrow a + (Y_{2i-1} - Y_{2i})^2/2. \tag{7}$$

A conservative variance estimate is then $\hat{\rho} \leftarrow \max\{a/N, \varepsilon\hat{\mu}\}$.

3. Set $N = \Upsilon \cdot \hat{\rho}/\hat{\mu}^2$, $a \leftarrow 0$. For $i = 1, ..., N$ do:
$a \leftarrow a + Y_i$.
The output is $\tilde{\mu} \leftarrow a/N$.

The $\mathcal{AA}$ algorithm uses a multistage approach to avoid using a priori information about $Y$. In the first step the algorithm uses the stopping rule theorem in its first step to generate a rough approximation of $\mu$. The failure probability is reduced to compensate for the multiple stage method. The second step estimates the variance of the samples, therefore the $\mathcal{AA}$ uses more information about the parameter of interest than the bounded variance which uses a uniform bounds. Finally, sequential analysis techniques use outcomes of the experiments in steps one and two to more accurately estimate the number of samples required for the $(\varepsilon, \delta)$ estimate of $\mu$.

The number of samples run by $\mathcal{AA}$ is within a small constant factor of the least number of samples required to guarantee that the output is within relative error $\varepsilon$ with probability at least $(1 - \delta)$ [Dagum et al., 1995]

### 4.3 IMPLEMENTATION OPTIMIZATIONS

There are a number of inefficiencies in the implementations of the bounded variance and $\mathcal{AA}$ algorithms as

$\mathcal{AA}$ **Step 3:**
$S_E \leftarrow 0$, $N_E \leftarrow \Upsilon \cdot \hat{\rho}_E/\phi_E^2$
for $i \leftarrow 1$ to $n$
$\quad N_i \leftarrow \Upsilon \cdot \hat{\rho}_i/\phi_i^2$, $S_i \leftarrow 0$

for $j \leftarrow 1$ to max$[N]$
$\quad$ **if** $j < N_E$ **then**
$\qquad z_E \leftarrow$ GenerateInstance$(Z, E)$
$\qquad \omega_E \leftarrow \prod_{j=1}^k \Pr[E_j = e_i | \pi(E_i)]|_{E = e, Z = z_E}$
$\qquad S_E \leftarrow S_E + \omega_E$

$\quad$ for $i \leftarrow 1$ to $n$
$\qquad$ **if** $j < N_i$ **then**
$\qquad\qquad z_{\mathcal{W}_i} \leftarrow$ Generate Instance $(Z \backslash X_i, \mathcal{W}_i)$
$\qquad\qquad \omega_i \leftarrow \prod_{j=1}^k \Pr[W_j = w_j | \pi(W_j)]|_{\mathcal{W}_i = w, Z = z_{\mathcal{W}_i}}$
$\qquad\qquad S_i \leftarrow S_i + \omega_i$

$\phi_E = S_E/N_E$
for $i \leftarrow 1$ to $n$
$\quad \phi_i \leftarrow S_i/N_i$
$\quad \hat{\mu}_i = \phi_i/\phi_E$

**Output:** $\hat{\mu}_i$, $i = 1, ..., n$

**Figure 3.** $\mathcal{AA}$ algorithm step 3

described. Most importantly, the pseudo-code implies that a new instance of the network, $z_{\mathcal{W}}$, must be generated for each query node, $X_i$, in each iteration of the algorithm. However, if a root query node is in set to its query state in the instance $z_E$ —the sample generated when scoring $\Pr[E = e]$ —we can simply rescore the network without generating a new instance. If the query node has parents and is in its query state in $z_E$ then its ancestors (direct and indirect) must be resampled before rescoring to ensure we score an independent sample. Since most nodes of interest are often root nodes, or have few parents, this can reduce the number of instances the algorithms generate. A node will of course be in it's state of interest with probability $\Pr[X_i = x_i | \pi(X_i)]$, or the prior probability for root nodes.

In the $\mathcal{AA}$ algorithm results of steps 1 and 2 are used to obtain an estimate on the number of samples required in step 3 to achieve a $(\varepsilon, \delta)$ estimate. The number of samples required to estimate the variance in step 2 may be close to the number of samples that will be required in the final step, so scoring of the query nodes can progress in steps 1 and 2, often avoiding the need for step 3.

### 4.4  STRATIFICATION TECHNIQUES

Consider a large network with $n$ query nodes for which we seek posterior probability estimates given evidence. In the worst case, bounded variance and $\mathcal{AA}$ will require $n + 1$ instances of the network to be generated each iteration. If the prior probabilities of the query nodes are small there are minimal savings from the rescoring method discussed in the previous section.

In practice we may not be interested in an accurate probability estimate for very improbable hypotheses. Bounded variance and $\mathcal{AA}$ can easily be modified to preferentially sample a subset of query nodes to reduce the number of generated instances.

A *stratification distribution* can be used to determine the frequency at which query nodes are instantiated and scored in each iteration of the algorithm. We define $f_i$ to be the frequency of instantiating $X_i = x_i$, and $\sum_{i=1}^{n} f_i = 1$. In every iteration we select at random from the stratification distribution to decide which query node $X_i$ we will score.

The selection of the stratification distribution depends on the goals of the belief network. If we are interested in the distribution of hypotheses then the frequency of selecting $X_i = x_i$ to sample can be proportional to its current marginal probability estimate. Initially the stratification distribution is uniform. A new stratification distribution is calculated at regular intervals, say 1000 iterations. In this scheme there is no overall reduction in the number of instances generated compared to the unmodified algorithm, but estimates for higher probability query nodes converge earlier.

If we want to find the most probable hypothesis then sampling can stop when it is unlikely that the probability of competing hypotheses is greater than the most probable query node. Assume an inference with two query nodes

$X_i$ and $X_j$ with runtime sampling estimates $\phi_i$ and $\phi_j$, and exact values $\rho_i$ and $\rho_j$. Assume that $\phi_i > \phi_j$. We wish to calculate the overall failure probability $\Pr[\rho_i < \rho_j]$. A point, $\rho_0$, is selected that lies between the estimates such that $\rho_0 = \phi_j(1 + \varepsilon) = \phi_i(1 - \varepsilon)$, thus $\varepsilon = (\phi_i - \phi_j)/(\phi_i + \phi_j)$. The upper bound on the failure probability is $\Pr[\rho_i < \rho_j] \leq \delta_i + \delta_j$

When the overall failure probability is less than some threshold we can be confident that $\phi_i > \phi_j$ and stop the algorithm. Finding the $m$ most probable diseases is a straightforward extension.

## 5  EVALUATION

We implemented the likelihood weighting, bounded variance and $\mathcal{AA}$ algorithms and tested them on a 146 node, multiply connected belief network from a medical domain. The network is amenable to exact inference, we used the Lauritzen and Spiegelhalter algorithm [Jensen et al., 1990; Lauritzen & Spiegelhalter, 1988] to provide the exact values for the experiments.

### 5.1  METHOD

Sixty five test cases were generated by sampling from the network. Findings were randomly selected for inclusion in each case to vary the number observations. The resulting range was from 1 to 91 observations, with a mean of 34.5. The cases generated for the evaluations varied in their difficulty, with over 10% of the cases having a probability of evidence $\Pr[E=e]$ in the order of $10^{-8}$ or smaller.

The bounded variance and $\mathcal{AA}$ algorithms were run with parameters $\varepsilon = 0.05$ and $\delta = 0.05$ —a 5% relative error with 5% failure probability.

We set an upper limit of 50,000 iterations for the algorithms. Note that for bounded variance and $\mathcal{AA}$ a single iteration may result in the generation of two instances (for one query node), where likelihood weighting will generate one instance per iteration. The running time was also measured. To compensate for the higher overhead in running the bounded variance and $\mathcal{AA}$ algorithms, we also compared the result of letting likelihood weighting run for 100,000 iterations. The different results are denoted LW50K and LW100K. In each case we measured the final posterior (marginal) probability on a disease node in the network, and calculated the relative error. The summary results are presented in Table 1. The algorithms are implemented in C and run on a 133MHz PowerPC 604 processor

### 5.2  RESULTS

The bounded variance and $\mathcal{AA}$ algorithms show considerable improvement over the traditional likelihood weighting algorithm, even when likelihood weighting was generating more instances on average. The convergence for likelihood weighting was poor, with the mean error reducing from 30.04% to 16.67%, however the percentage of estimates which had greater than 5% error only reduced slightly from 79.37% to 73.02%.

In some cases there was greater than 5% relative error in both bounded variance and $\mathcal{A}\mathcal{A}$. Interestingly, all but one of these cases for the new algorithms occurred when less than 30% of the recommended iterations had been completed when the designated maximum number of iterations had been reached. The relative error can be calculated at any given stage of the simulation. This information was not available for likelihood weighting.

In the majority of cases the $\mathcal{A}\mathcal{A}$ algorithm generated fewer instances than the bounded variance algorithm, consequently the bounded variance algorithm was slightly more accurate. In some cases with extremely rare evidence bounded variance stopped before $\mathcal{A}\mathcal{A}$, this occurred because in step 2 the $\mathcal{A}\mathcal{A}$ algorithm uses a conservative estimate of the sample variance.

A Tukey test of multiple comparison with a family error rate of 0.05 resulted in statistically significant differences between $\mathcal{A}\mathcal{A}$ and the likelihood weighting algorithms, and also between bounded variance and likelihood weighting. There was no statistically significant difference between the $\mathcal{A}\mathcal{A}$ and bounded variance algorithms in this multiple comparison.

# 6   DISCUSSION

The results presented in the previous section show that the bounded variance and $\mathcal{A}\mathcal{A}$ algorithms perform significantly better than straight likelihood weighting. The new algorithms provide a larger class of networks amenable to polynomial time approximation compared to existing Monte Carlo sampling techniques. By improving the method of weighting samples we can take advantage of recent advances in stopping rules for Monte Carlo sampling to provide run-time information on convergence.

The cost for these improvements in performance is the increase in computational effort. The $\mathcal{A}\mathcal{A}$ and bounded variance algorithms may be forced to generate a new instance for each hypothesis node being queried. Note that instance generation is linear in the number of nodes of a network. In our example we were only interested in the marginal probability of one hypothesis node so the overhead is small. For larger networks with many nodes of interest the constant factor overhead may be a reasonable trade-off to potentially exponential approximation times with other sampling algorithms. We presented a variety of adaptive stratification strategies to improve the running time of the algorithms in specific situations where a tight estimates are not required on all query nodes. This technique is complementary to importance sampling. Since the stratified approach sets each hypothesis node in turn we avoid the problem of making the combination of rare hypotheses more likely than they should be, which may occur in importance sampling.

We know that techniques such as importance sampling [Shachter & Peot, 1990] and Markov blanket scoring can improve convergence rates of the likelihood weighting algorithm [Cousins et al., 1993; Shwe & Cooper, 1991]. These variance reduction techniques may also be applied to the new algorithms to improve their convergence rates, while retaining their other advantages. The incremental variance estimation technique used in the AA algorithm (Eq. 7) can easily be used to measure whether importance distributions do result in decreased variance.

## 6.1   FUTURE WORK

We plan to implement importance sampling for the new algorithms, and to test them on larger networks. The combination of importance sampling with stratification should lead to considerable improvements in efficiency since hypothesis nodes of interest will be sampled in their query state with higher probability. We are also exploring utility-directed applications of these algorithms.

The AA algorithm's efficiency can be improved. In the current implementation the number of samples to estimate the variance in step 2 can be large when the parameter of estimation is small. However, we are only interested to see if the sampled variance is smaller than $\varepsilon\mu$. Numerous sequential methods [Siegmund, 1985] may allow us to detect this with confidence well before completing the recommended number of samples.

**Table 1:** Results of inference algorithm comparison on 65 test cases in a 146 node belief network for the $\mathcal{A}\mathcal{A}$ algorithm, bounded variance (BV), 50,000 (LW50K) and 100,000 (LW100K) iterations of likelihood weighting.

|  | AA | BV | LW50K | LW100K |
|---|---|---|---|---|
| Mean relative error | 2.78% | 2.13% | 30.04% | 16.67% |
| Std.dev. relative error | 7.23% | 5.44% | 49.75% | 16.87% |
| Mean instances generated | 77,339 | 80,562 | 50,000 | 100,000 |
| Mean time (seconds) | 28.12 | 28.95 | 21.62 | 40.95 |
| Cases with > 5% error | 9.52% | 11.11% | 79.37% | 73.02% |
| Cases with > 5% error and > 30% completion | 1.54% | 1.54% | NA | NA |

**Acknowledgments**

Our thanks to Ross Shachter for ideas on stratified sampling. Comments from the anonymous reviewers helped to improve the presentation of this paper. This work was supported by the National Science Foundation under grant IRI-93-11950, and by the Stanford University CAMIS project under grant IP41LM05305 from the National Library of Medicine of the National Institutes of Health.